\documentclass[letterpaper, 10 pt, conference]{ieeeconf}  %

\IEEEoverridecommandlockouts                              %

\usepackage{amsmath}%
\usepackage{amssymb}%
\usepackage[dvipsnames]{xcolor}%
\newcommand{\gc}[1]{\textcolor{gray}{#1}}
\usepackage{bm}%
\usepackage{graphicx}
\usepackage{booktabs}

\usepackage{tikz}

\usepackage{amsthm}
\theoremstyle{plain}
\newtheorem{definition}{Definition}

\newtheorem{problem}{Problem}
\usepackage{hyperref}
\hypersetup{%
  colorlinks=true,%
  linkcolor={red!50!black},
  citecolor={blue!65!black},
  urlcolor={blue!80!black},
  bookmarksnumbered=true,%
  bookmarksopen=true}
\usepackage[capitalise]{cleveref}
\usepackage{caption}
\makeatletter
\let\oldsine\sin
\let\oldcos\cos

\renewcommand{\sin}{\@ifnextchar\bgroup\sin@arg\oldsine}
\renewcommand{\cos}{\@ifnextchar\bgroup\cos@arg\oldcos}

\newcommand{\sin@arg}[1]{s_{#1}}
\newcommand{\cos@arg}[1]{c_{#1}}
\makeatother

\title{\LARGE \bf
Fast and Robust Temporal Logic Planning\\via ADMM-based Trajectory Optimization}

\author{Lukas Pries$^{1,*}$, Joris Verhagen$^{2,*}$, Jon Arrizabalaga$^{3}$, Jana Tumova$^{2}$, Markus Ryll$^{1}$, Zachary Manchester$^{3}$%
\thanks{$^{*}$ Contributed equally to this work}
\thanks{$^{1}$Department of Aerospace and Geodesy, TU Munich, Germany,
{\tt\small \{lukas.pries,markus.ryll\}@tum.de}}%
\thanks{$^{2}$Division of Robotics, Perception, and Learning, KTH Royal Institute of Technology, Stockholm, Sweden
        {\tt\small \{jorisv, tumova\}@kth.se}}%
\thanks{$^{3}$Department of Aeronautics and Astronautics, MIT, Cambridge, MA, USA
        {\tt\small \{jonarri, zacm\}@mit.edu}}%
}

\begin{document}

\maketitle

\thispagestyle{empty}
\pagestyle{empty}

\begin{abstract}

We present a fast numerical method for safe continuous-time motion planning under Temporal Logic (TL) specifications.
The method generates smooth continuous trajectories that remain collision-free while robustly satisfying temporal and logical task requirements.
A central component of our method is the formulation of nonconvex safety and logic constraints as unions of convex sets where associated discrete decisions are encoded in a joint feasibility graph.
This graph representation allows Euclidean projection onto the feasible set and proximal robustness maximization to be reformulated as shortest- and widest-path problems, respectively.
Building on this structure, we develop a nonconvex splitting method based on the Alternating Direction Method of Multipliers (ADMM), which decouples smooth spatio-temporal trajectory optimization from nonsmooth discrete constraint handling within the optimization.
The resulting algorithm exhibits reliable convergence across benchmarks and scales to large-scale motion-planning problems, providing a $4.7\times$ average speedup over the state of the art on discrete and continuous-time logic problems.
\end{abstract}

\section{INTRODUCTION}
Motion planning in real-world environments requires reasoning about both continuous and discrete decisions.
Discreteness arises predominantly from two sources: (i) obstacles forcing a robot to choose from one of multiple possible collision-free paths, and (ii) task specifications creating ambiguity in how they can be satisfied.
Prior works address (i) via sampling-based planning~\cite{orthey2023sampling}, trajectory optimization~\cite{schulman2013finding}, or mixed-integer programming (MIP)~\cite{deits2015efficient} which can be extended to also involve (ii) via automaton progress for temporal logic specifications~\cite{karlsson2020sampling,vasile2020reactive}, smooth $\min$ and $\max$ operators~\cite{gilpin2020smooth}, or mixed-integer constraints~\cite{raman2014model}.
\\
This work focuses on the problem from the perspective of mixed-integer programming and automata progress.
Both obstacle avoidance and temporal logic specifications admit a natural representation as unions of convex sets: each set encodes a continuous subproblem and the discrete problem reduces to selecting which sets to traverse.
\\
Recent work has shown that this problem can reliably be solved via convex relaxations on Graphs of Convex Sets (GCS)~\cite{marcucci2024shortest}, where the discrete decisions are relaxed yet empirically yield globally optimal solutions.
Extensions handle temporal logic via product graphs of the GCS and the automaton~\cite{kurtz2023temporal,chen2026signal}, but the resulting optimization problems are expensive to solve, do not remain convex under common actuation and continuity constraints, and cannot maximize robust satisfaction of the specification.
Two-stage approaches solve either a discrete problem to satisfy temporal logic and then find a continuous trajectory~\cite{fainekos2005temporal,da2021automatic}, or solve a continuous but simplified trajectory problem over the specification which is then refined to tackle additional constraints~\cite{ren2025accelerating}.
\\
We instead propose a \emph{splitting} approach, separating continuous and discrete decision-making within a joint framework via the Alternating Direction Method of Multipliers (ADMM)~\cite{takapoui2020simple}.
Splitting the smooth continuous optimization from the search over the workspace- and temporal logic decision graph removes heuristic restrictions of decoupled approaches and enables parallelized computation, more general constraint handling and explicit robustness maximization.

\begin{figure}[t!]
    \centering
    \includegraphics{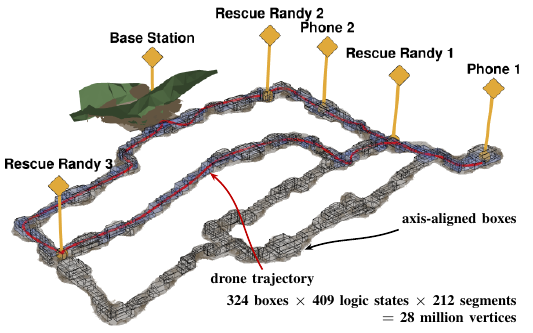}
    \vspace{-5mm}
    \caption{Our ADMM solver on an underground cave-rescue scenario. A temporal logic specification requires visiting the rescue Randys and phones as specified in \cref{sec:validation}. The total product graph of the Transition- and Logic Graph has 28 million vertices, beyond the capacity of existing methods. The final trajectory is shown in red and visited boxes shaded blue.}
    \label{fig:1}
    \vspace{-5mm}
\end{figure}

\subsection*{Contributions}
The main contributions of this work are:

\begin{itemize}

\item \textbf{A graph-based formulation.} We represent collision-avoidance and Signal Temporal Logic specifications as \emph{unions of convex sets} encoded jointly into a \emph{joint feasibility graph}. This structure allows Euclidean projection onto the feasible set and proximal robustness maximization as \emph{shortest- and widest-path problems}.

\item \textbf{An efficient splitting method for spatio-temporal optimization.} Building on this, we develop an \emph{ADMM-based solver} separating smooth nonlinear optimization over continuous spatial and temporal variables from discrete and nonsmooth feasibility and robustness operations, yielding \emph{efficiently solvable subproblems}.

\end{itemize}

\clearpage
\section{PRELIMINARIES}

\subsection{Spatio-Temporal Trajectory Optimization}
\label{sec:spatiotempoptim}

Consider a spatio-temporal objective that jointly minimizes trajectory smoothness and duration time,
\begin{equation}
\label{eq:st_obj}
\min_{q,T}
J(q,T)
=
\min_{q,T}\int_0^T
\sum_{r=1}^{s} w_r
\big\|q^{(r)}(t)\big\|_2^2\,\mathrm dt
+ w_T T,
\end{equation}
where $r$ indicates derivative order, $w_r\geq0$, and $w_T>0$. Restricting the trajectory $q: [0,T]\rightarrow\mathbb{R}^d$ to piecewise polynomials renders \eqref{eq:st_obj} finite-dimensional.
Let $q$ consist of $M$ polynomial segments of degree $N=2s-1$ with durations $T_m>0$, such that $T=\sum_mT_m$. Each segment $q_m$ admits two equivalent bijective parameterizations: a B\'ezier representation that facilitates the enforcement of convex continuous-time constraints and a knot-state representation that exposes an efficient structure for minimizing the objective.

\paragraph{B\'ezier Representation}
\label{sec:bezier}
Using normalized time $\tau=t/T_m\in[0,1]$, segment $m$ can be expressed as
\begin{equation}
q_m(T_m\tau)
=
\sum\nolimits_{j=0}^{N} B_j^N(\tau) \, b_{m,j},
\end{equation}
where $b_{m,j}\in\mathbb{R}^d$ are B\'ezier control points.
This representation separates the spatial variables $\mathbf{b}_m = (b_{m,j})_{j=0}^N$ from the duration $T_m$ and provides finite-dimensional convex certificates for continuous-time constraints through its convex-hull property: Consider a convex set $\mathcal P\subseteq\mathbb{R}^d$, requiring $b_{m,j}\in\mathcal P$
for all control points guarantees $q_m\in\mathcal P$ over the complete segment. The trajectory is parameterized by $x = (\mathbf{b}, \mathbf{T})$ with $\mathbf{b} = (\mathbf{b}_m)_{m=1}^M\in\mathbb{R}^{M(N+1)\times d}$ and $\mathbf{T} = (T_m)_{m=1}^M\in\mathbb{R}^M_{>0}$.

\newcommand{\xl}{\tilde{x}}

\paragraph{Derivatives} \label{sec:derivatives}
Derivatives are again B\'ezier curves with their control points defined recursively through linear combinations $b^{(r)}_{m,j} = (N-r+1) \big(b^{(r-1)}_{m,j+1} - b^{(r-1)}_{m,j}\big)$, yielding
\begin{equation}
\label{eq:bezier_derivative}
q_m^{(r)}(T_m\tau)
=
\tfrac{1}{T_m^r}
\sum\nolimits_{j=0}^{N-r} B_j^{N-r}(\tau)\,b^{(r)}_{m,j}.
\end{equation}
For a convex admissible set $\mathcal P_r\subseteq\mathbb{R}^d$ for the $r$-th derivative, the constraint
$q_m^{(r)}(t)\in\mathcal P_r$ is therefore sufficiently enforced by
\begin{equation}
\label{eq:bezier_derivative_constraint}
b^{(r)}_{m,j} \in \tilde{T}_{m,r}\mathcal P_r \quad \forall j,
\end{equation}
where each $b^{(r)}_{m,j}$ is linear in the original control points and $\tilde{T}_{m,r} = T_m^r$ represents a time-dependent scaling parameter. \eqref{eq:bezier_derivative_constraint} is a perspective constraint of the convex set $\mathcal P_r$ and therefore jointly convex in the spatial variables $\mathbf{b}_m$ and the \emph{lifted temporal scaling variables} $\mathbf{\tilde{T}}_{m}(T_m) = (1, T_m, \dots, T_m^N)$.
Bounds on higher-order derivatives can thus be formulated as continuous-time convex constraints in the lifted variables $\xl = g(x) = (\mathbf{b}, \mathbf{\tilde{T}}(\mathbf{T}))$, for which a smooth mapping exists.

\paragraph{Knot-State Representation}
Each segment is equivalently represented by its derivative states $\mathbf{d}_m = \big(q_m(T_m), \ldots, q_m^{(s-1)}(T_m) \big)\in\mathbb{R}^{s\times d}$ at its two endpoints. The bijective mapping is defined by
$
\begin{bmatrix}
\mathbf{d}_{m-1} \ \mathbf{d}_{m}
\end{bmatrix}
=
A(T_m)\mathbf{b}_m
$
and $A(T_m)$ is nonsingular for every $T_m>0$~\cite{wang2020alternating}.
Sharing the knot state between adjacent segments enforces $C^{s-1}$ continuity by construction.
Consequently, the trajectory can equivalently be parameterized by the knot states $\mathbf{D}=(\mathbf{d}_m)_{m=0}^M$ and the durations $\mathbf T$.
In these variables, \eqref{eq:st_obj} is unconstrained and amenable to alternating minimization~\cite{wang2020alternating}.
\subsection{Temporal Logic}
\label{ssec:preliminaries_tl}

We use Signal Temporal Logic (STL)~\cite{donze2010robust} to specify spatial and temporal requirements on continuous robot trajectories. For example, the robot should \emph{always eventually charge} or \emph{be near one of the WiFi-routers for a certain duration} to transmit data.
We can define specifications w.r.t. the satisfaction of a concave function $\mu$ over $q(t)$, or observation of \emph{atomic predicates}, $\mathsf{AP}$
\begin{equation}
\label{eq:stl_predicate}
    p_i : \mu_i(q(t)) \geq 0, \qquad \mu_i:\mathbb{R}^d\rightarrow \mathbb{R}
\end{equation}
which induces a labeling along the trajectory
\begin{equation}
\label{eq:stl_labelling}
    \mathrm{Lab}(q, t) = \{p_i \in \mathsf{AP}: \mu_i(q(t)) \geq 0 \}.
\end{equation}
Examples of $\mathsf{AP}$s are then \emph{Charging region} or \emph{WiFi-reception}.
Let $I=[t_1,t_2] \subseteq \mathbb{R}_{\geq 0}, t_2 > t_1$ be a closed and bounded interval.
Our fragment of STL is defined as
\begin{align}
\label{eq:stl_fragment}
  \psi     &:= \top \mid p \mid \neg p \mid \psi_1 \land \psi_2 \mid \psi_1 \lor \psi_2, \\
  \phi     &:= F_I\psi
            \mid G_I\psi
            \mid \psi_1 U_I \psi_2 \notag \mid F_{I_1}G_{I_2}\psi
            \mid \phi_1 \land \phi_2
            \mid \phi_1 \lor \phi_2 . \notag
\end{align}
with $p \in \mathsf{AP}$ and $F_I$, $G_I$, and $U_I$ are the \emph{eventually}, \emph{always}, and \emph{until} operators.
A trajectory $q$ at time $t$ satisfies these predicates and operators according to
\begin{equation}
    \label{eq:stl_semantics}
    \begin{alignedat}{2}
        &q,t \models p &&\quad\iff \mu(q(t))\geq 0 \\
        &q,t \models \neg p &&\quad\iff \mu(q(t)) < 0\\
        &q,t \models \phi_1 \land \phi_2 &&\quad\iff q,t \models \phi_1 \ \text{and} \ q,t\models\phi_2 \\
        &q,t \models \phi_1 \lor \phi_2 &&\quad\iff q,t \models \phi_1 \ \text{or} \ q,t\models \phi_2 \\
        &q,t \models F_I \psi &&\quad\iff \exists \tau \in t+I: q,\tau \models \psi \\
        &q,t \models G_I \psi &&\quad\iff \forall \tau \in t+I: q,\tau \models \psi \\
        &q,t \models \psi_1 \mathcal{U}_I\psi_2 &&\quad\iff \begin{aligned}[t] &\exists \tau \in t+I: q,\tau \models \psi_2, \\ &q,s \models\psi_1 \quad \forall s \in [t,\tau). \end{aligned} \\
        &q,t \models F_{I_1}G_{I_2}\psi &&\quad\iff \exists \tau \in t+I_1: q,\tau \models G_{I_2} \psi
    \end{alignedat}
\end{equation}

\begin{figure}[t]
\centering

\includegraphics{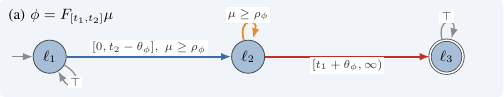}

\vspace{0.25em}

\includegraphics{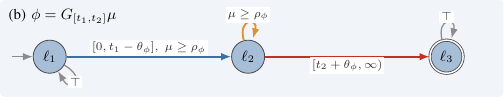}

\vspace{0.25em}

\includegraphics{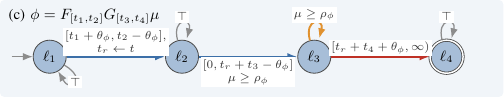}

\vspace{0.25em}

\includegraphics{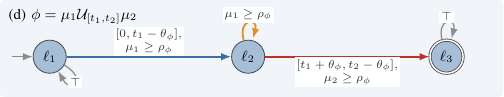}

\caption{
\textbf{Temporal operator automata.}
Timed automata for the \emph{Eventually}, \emph{Always}, \emph{Eventually Always} and \emph{Until} operators.
The array $[t_1,t_2]$ denotes the interval in which the global time $t$ must reside.
Spatial robustification is achieved by robustifying the predicate satisfaction via $p_i^{\rho}: \mu(\cdot) \geq \rho_{\phi}$ and temporal robustification is achieved by extending the intervals.
$F_{[t_1,t_2]}G_{[t_3,t_4]}\mu$ is robustified by recording the reset time, $t_r \gets t$, in the center of $[t_1,t_2]$ and observing $\mu$ ``$\theta_{\phi}$" longer and shorter than required by $[t_3,t_4]$.
Similarly, $\phi_1 \mathcal{U}_{[t_1,t_2]}\phi_2$ is robustified by observing $\mu_1$ earlier than necessary, and switching to $\mu_2$ in the middle of the interval $I=[t_1,t_2]$.
}
\label{fig:temporal-operator-automata}
\vspace{-3mm}
\end{figure}

The boolean semantics permit a timed-automaton representation, inducing a word given a sequence of labels and the times at which these labels occur
\begin{definition}[Timed Automaton]
\label{def:timed-automaton}
The \emph{timed automaton} of specification $\phi$ is the tuple $\mathcal{A}_\phi = (L, \ell_0, L_F, C, \Delta)$, where $L$ is a finite set of locations, $\ell_0$ the initial location, $L_F \subseteq L$ the accepting locations, $C$ a finite set of clocks, and
\begin{equation}
  \Delta \subseteq L \times 2^{\mathsf{AP}} \times \Gamma(C)
          \times 2^{C} \times L
  \label{eq:ta-transitions}
\end{equation}
the set of transitions.
A transition $(\ell, \lambda, \gamma, \varrho, \ell') \in \Delta$ is enabled when the current predicate labeling satisfies $\mathrm{Lab}(q, t) = \lambda$ and the clock valuation satisfies the guard $\gamma \in \Gamma(C)$.
The transition moves the automaton from $\ell$ to $\ell'$, resets the clocks in $\varrho \subseteq C$, progressing the formula accordingly.
\end{definition}

\subsection{Robust STL}
In many scenarios, maximizing robustness of the specification w.r.t. spatial and temporal perturbations can be important.
For example, a robot with an uncertain state estimate may want to robustly, rather than nominally, satisfy the predicate function $\mu_i(\cdot)\geq0$ and it may want to satisfy it for a longer duration than strictly necessary. Two traditional examples of robustness metrics are spatial and temporal robustness.
\emph{Spatial robustness} is defined for a predicate $p_i$ as $\rho_{p_i}(q,t) = \mu_i(q(t))$. Larger values indicate higher robustness w.r.t. signal perturbations.
\emph{Temporal robustness} $\theta_{p_i}(q,t)$ is intuitively the maximal temporal shift of the signal to the past or future at $t$ without violating $p_i$.
The values of spatial and temporal robustness $\rho_{\phi}$ and $\theta_{\phi}$ for a general formula $\phi$ are defined recursively from here using min and max operators \cite{donze2010robust}. %
We denote $\delta_\phi\in\{\rho_\phi,\theta_\phi\}$ as a robustness metric. %

\newcommand{\pset}{\mathcal{P}}
\newcommand{\lset}{\mathcal{L}}
\newcommand{\sset}{\mathcal{S}}
\newcommand{\tset}{\mathcal{T}}
\newcommand{\xset}{\mathcal{X}}

\newcommand{\xref}{x'}

\section{PROBLEM STATEMENT}
We consider the robust STL motion-planning problem for a robot operating under temporal logic tasks and kinodynamic constraints in a cluttered environment.
We do not impose the full system dynamics explicitly but require the trajectory and its derivatives to be smooth and sufficiently bounded to be compatible with differentially flat systems, a broad class of controllable linear and nonlinear systems.

\newcommand{\vv}{q}
\begin{problem}[Robust STL Motion Planning]
\label{prob:general}
\begin{subequations} \label{eq:1problem}
\begin{align}
    \min_{\vv, T}& & & J(q,T) - w\,\delta_\phi(q,0)& \text{(objective)}\\
    \text{s.t.}
    &&& \vv \in \mathcal{D}, & \text{(dynamics)} \label{eq:1feasibility} \\
    &&& \vv \in \sset, & \text{(collision-free)}\label{eq:1safety} \\
    &&& \delta_{\phi}(q, 0) \geq 0. & \text{(STL satisfaction)}
\end{align}
\end{subequations}
\end{problem}
Problem 1 is difficult for two compounding reasons.
First, the feasible set is highly nonconvex: $\mathcal{S}$ and $\phi$, via $\mathcal{A}_{\phi}$, require discrete choices along the trajectory, e.g., going left or right of each obstacle and selecting the predicates and timestamps to satisfy the specification.
Second, these choices must be mutually consistent: a geometric branch determines which logical assignments remain reachable in time, so the two cannot be resolved separately.
In addition, the objective is nonsmooth, since $\delta_\phi$ quantifies the robustness over $\phi$.

\begin{figure*}[t]
    \centering
    \includegraphics{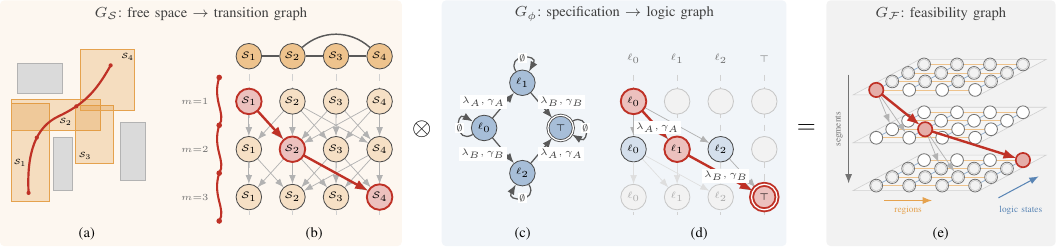}
    \vspace{-5mm}
    \caption{\textbf{Graph representations.}
    (a) The free space is covered by overlapping convex regions $\mathcal{S}_o$ and the trajectory is split into segments $m=1,\dots,M$. (b) The possible assignments of segments to adjacent regions form the transition graph $G_\mathcal{S}$. (c) The STL specification $\phi$ is compiled into a timed automaton and (d) unrolled over the same segments into the logic graph $G_\phi$. (e) Their product $G=G_\mathcal{S}\otimes G_\phi$ is the feasibility graph: each layer holds one vertex per region--logic-state pair, so a path (red) from the initial vertex to an accepting one is a segment-to-region assignment that jointly satisfies $\sset$ and $\phi$.}
    \label{fig:graphs}
    \vspace{-5mm}
\end{figure*}

\section{Graph Representations of the Decision Spaces}
\label{sec:graphs}
We first show how the non-convexity of collision-avoidance and temporal logic constraints can be captured by the same machinery: a graph construction from a union of convex sets, either from convex space composition or from convex predicate and clock constraints.
The joint problem amounts to finding a \emph{path} through the product of the respective \emph{Transition} and \emph{Specification Logic Graph}, the robustified \emph{Feasibility Graph}, and the trajectory residing within the associated convex constraints.

\subsection{Transition Graph}
We assume the free space is given as a finite union of closed convex regions $\bigcup_{o \in \mathcal{O}} \sset_o$, e.g. a set of bounding boxes, ellipses or polytopes,
as produced by region inflation methods such as \textsc{Iris-ZO}~\cite{werner2024faster}.
Confining segment $m$ to region $\sset_o$ is a convex constraint on the trajectory parameters
\begin{equation}
\sset_{m,o} := \{\, x=(\mathbf{b}, \mathbf{T}) \,|\, b_{m,j} \in \mathcal{S}_{o}, \,\forall j\},
\end{equation}
by the convex-hull property of the B\'ezier representation
(\ref{sec:spatiotempoptim}). Collision avoidance along the trajectory therefore requires every segment to lie in at least one safe region.
Following \cite{pries2026admm}, we build a graph $G_\sset$ over segment-region assignments in which paths correspond to the admissible choices. Edges are included only between intersecting regions, so that consecutive segments are assigned to overlapping regions.
\begin{definition}[Transition Graph]
\label{def:transition-graph}
The transition graph $G_\sset = (V_\sset, E_\sset)$ contains one vertex $v_{m,o}^\sset$ for every segment $m \in \mathcal M$ and region $o \in \mathcal{O}$. An edge connects $v_{m,o}^\sset$ to $v_{m+1, o'}^\sset$ if $\sset_{o} \cap \sset_{o'} \neq \emptyset$.
Every vertex $v_{m,o}^\sset$ carries the convex constraint set
$\sset_{m, o}$, which confines segment
$m$ to region $o$.
\end{definition}

By construction, $G_\sset$ is acyclic and layered by the segment index: every edge leads from layer $m$ to layer $m+1$ as depicted in Fig. \ref{fig:graphs}.
A \emph{path}, written $\pi_\sset \in G_\sset$ for brevity, is a sequence
$\pi_\sset = (v_{1,o_1}^\sset, \dots, v_{M, o_M}^\sset)$
of vertices which induces the collision-free set $\sset_{\pi_\sset} = \bigcap_{m\in\mathcal{M}} \sset_{m,o_m}$. The union over all admissible paths yields,
\begin{equation}
    \sset = \bigcup_{\pi_\sset \in G_\sset} \sset_{\pi_\sset}.
\end{equation}

\subsection{Robustification of Timed Automata}
To maximize the robustness over $\phi$ in the objective of Problem~\ref{prob:general}, yet still model the decision space as a union of convex sets, we propose spatially and temporally robustifying these sets.
Starting from the TA of $\phi$, $\mathcal{A}_{\phi}$, we first show how it can be robustified to $\mathcal{A}_{\phi}^{\delta}$.

\paragraph{Spatial Robustness}
Every predicate guard $\mathrm{Lab}(q,t)=\lambda$ in the TA is robustified with a scalar margin $\rho_{\phi}$ by enabling transitions with condition $\mathrm{Lab}(q,t)=\lambda_{\rho_\phi}$ with $\lambda_{\rho_\phi}:=\{p_i \in \lambda: \mu_i(q(t)) \geq \rho_{\phi}\}$ where $\lambda_{\rho_\phi}\subseteq\lambda$.
This procedure is shown in \cref{fig:temporal-operator-automata}.

\paragraph{Temporal Robustness}
Every clock guard $\gamma \in \Gamma(C)$ is a conjunction of bounds $t_{c,1} \leq t \leq t_{c,2}$ on clock $c \in C$. The temporal robustness $\theta_{\phi}$ cannot be applied uniformly to these bounds but must instead \emph{contract} or \emph{expand} each one, depending on the operator in \cref{eq:stl_fragment}; e.g., a guard tightened on both sides is robustified as $\gamma_{\theta_\phi}:=\{t \,|\, t_{c,1} + \theta_{\phi} \leq t \leq t_{c,2} - \theta_{\phi}\}$ with $\gamma_{\theta_\phi}\subseteq\gamma, \,\forall \gamma \in \Gamma(C)$.
Rather than formally defining the temporal robustification for each operator, we refer to the robustified operators in \cref{fig:temporal-operator-automata}.

We note that conjunctions and disjunctions of these operators preserve this construction via the synchronous product and union.
Negation is covered by shrinking or extending visiting durations of regions without overlap with the negated predicate region.

\subsection{Specification Logic Graph}
The robustified TA of $\phi$, $\mathcal{A}_{\phi}^{\delta}$ enables transitions upon satisfaction of robustified guards.
Rolling out all possible traversals of $\mathcal{A}_{\phi}^{\delta}$ over the segments permits a set-based/constraint perspective where each segment $m$ is subject to robustified spatial- and temporal constraints.
First, for a clock \(c\in C\), let \(r_c(m)\) denote the index of the most recent transition before \(m\) that reset \(c\). Then, for a path in the automaton, every clock value $t_{m,c}(\mathbf T) = \sum_{i=r_c(m)+1}^{m} T_i$ is linear in \(\mathbf T\).
\newcommand{\gammai}{\gamma}
\newcommand{\lambdai}{\lambda}
More generally, temporal guards can be represented as $\gammai: a_{\gammai} t_{m,c}(\mathbf T) \geq b_{\gammai}$.
Hence, for each segment \(m\in\mathcal M\), the label $\lambdai$ and temporal guard $\gammai$ admit (robust) convex constraints on the trajectory parameters:
\begin{align}
    \lset_{m, \lambdai}(\rho_\phi) &:= \{x=(\mathbf{b}, \mathbf{T}) \,|\, \mu_i(b_{m,j}) \geq \rho_{\phi},\, \forall p_i \in \lambdai, \, \forall j\}, \nonumber \\
    \tset_{m, \gammai}(\theta_\phi) &:= \{x=(\mathbf{b}, \mathbf{T})\,|\, a_{\gammai} t_{m,c}(\mathbf{T}) \geq b_{\gammai} + \theta_{\phi}\}.
\end{align}
We define $G_{\phi}$ as the logic graph of specification $\phi$ and its TA $\mathcal{A}_{\phi}$
\begin{definition}[Logic Graph]
\label{def:logic-graph}
    In the \emph{logic graph} $G_\phi = (V_\phi, E_\phi)$,
    a vertex $v^\phi_{m,\lambdai,\gammai}$ represents the commitment of segment $m$ to the label $\lambdai$ and clock constraints $\gammai$ and carries convex constraint sets $\lset_{m,\lambdai}$ and $\tset_{m,\gammai}$, respectively.
    An edge connects $v^\phi_{m,\lambdai,\gammai}$ to $v^\phi_{m+1,\lambdai',\gammai'}$ if a vertex $l$ exists between the corresponding edges in $\mathcal{A}_{\phi}$;
    whenever applying label $\lambdai$ with clock guards $\gammai$ at $\ell$ progresses the specification to $\ell'$, that is,
    $e_m = (\ell,\lambdai,\gammai,\varrho,\ell') \in \Delta$ from \cref{def:timed-automaton}.
\end{definition}

An accepting path $\pi_\phi = (v_{1,\lambda_1,\gamma_1}^\phi,\dots,v^\phi_{M,\lambda_M,\gamma_M})$ through $G_{\phi}$ is a sequence of convex constraints, $((\lset_{1,\lambda_1},\tset_{1,\gamma_1}),\dots,(\lset_{M,\lambda_M},\tset_{M,\gamma_M}))$ and induces itself convex constraint sets as intersections $\lset_{\pi_\phi} = \bigcap_{m \in \mathcal{M}} \lset_{m,\lambda_m}$ and $\tset_{\pi_\phi} = \bigcap_{m \in \mathcal{M}} \tset_{m,\gamma_m}$.
The complete temporal-logic feasible set is obtained by taking the union over all accepting paths, written $\pi_\phi \in G_\phi$,
\begin{equation}
\lset_{\phi}(\delta_\phi) = \bigcup_{\pi_\phi \in G_\phi} \lset_{\pi_\phi}(\rho_\phi) \cap \tset_{\pi_\phi}(\theta_\phi),
  \label{eq:union_safety}
\end{equation}
which is, according to \cref{eq:1problem}, formulated via non-negative robustness maximization of $\delta_{\phi}\in\{\rho_\phi,\theta_\phi\}$.

Note that the spatial constraints are entirely separable for each segment $m$ as the rollout over $\mathcal{A}_{\phi}^{\delta}$ already specifies the temporal relationships between segments that permit satisfaction of $\phi$.
However, the temporal constraints for segment $m$ are a function of preceding segment durations.

\newcommand{\fset}{\mathcal{F}}

\subsection{Feasibility graph}
\label{sec:product}
The two independent \emph{transition} and \emph{logic} graphs should jointly restrict a feasible trajectory to sequence through collision-free regions and satisfy the right predicates at the right time.
We achieve a unified graph for the feasible set $\fset = (\sset \cap \lset_\phi) \cap \tset_\phi$ as the product $G_\fset = G_\sset \otimes G_\phi$
\begin{definition}[Feasibility Graph]
    The product vertex is denoted by $v_{m,o,\lambda,\gamma}^\fset = \big(v_{m,o}^\sset, v_{m,\lambda,\gamma}^\phi \big)$ and carries the convex constraint sets $\fset_{m, o, \lambda, \gamma} = \sset_{m,o} \cap \lset_{m,\lambda}$ and $\tset_{m,\gamma}$.
    Two vertices $v_{m,o,\lambda,\gamma}^\fset$ and $v_{m+1,o',\lambda',\gamma'}^\fset$ are connected if both component transitions are admissible,
    \begin{equation}
        \bigl(v_{m,o}^\sset,v_{m+1,o'}^\sset\bigr)\in E_\sset,
        \quad
        \bigl(v_{m,\lambda,\gamma}^\phi,v_{m+1,\lambda',\gamma'}^\phi\bigr)\in E_\phi. \nonumber
    \end{equation}
\end{definition}
A path $\pi$ selects one joint safety–logic decision for each segment and induces (robust) spatial and temporal feasibility sets $\fset_\pi(\rho_\phi)=\sset_{\pi}\,\cap\,\lset_{\pi}(\rho_\phi)$ and $\tset_{\pi}(\theta_\phi)$, respectively. The union over all paths $\pi \in G_\fset$ yields a union of convex sets,
\begin{equation}
    \fset(\delta_\phi) = \bigcup_{\pi \in G_\fset} \fset_\pi(\rho_\phi) \cap \tset_\pi(\theta_\phi),\quad \delta_{\phi}\in\{\rho_\phi,\theta_\phi\}.
\end{equation}

\section{Feasibility and Robustness Operators}
The graph representations of Sec.~\ref{sec:graphs} express the nonconvex feasible sets as unions of convex sets indexed by discrete path choices. This confines the nonconvexity to path selection, while each fixed path induces a convex projection problem. Exploiting this structure, we formulate feasibility projection and robustness maximization as shortest-path and widest-path problems, respectively.

\subsection{Convex Operators}
\label{sec}
We first introduce the convex primitives underlying both graph-search formulations.
Let $\xset\subseteq\mathbb{R}^{\tilde{N}}$ be a nonempty closed convex set.
The Euclidean projection of a reference point $\xref$ onto $\xset$ is defined by
\begin{align}
\Pi_{\xset}(\xref)
:=
\underset{x\in\xset}{\arg\min}
\frac{1}{2}\|x-\xref\|_2^2 .
\label{eq:convexproj}
\end{align}
Since $\xset$ is convex and the objective is strictly convex, \eqref{eq:convexproj} is a convex program with a unique solution.

We further consider a margin-parameterized set $\xset(\delta)$ and its convex lifted feasible set
$
\widehat{\xset}
:=
\{(x,\delta)\mid x\in\xset(\delta)\}.
$
The corresponding robustness-augmented operator is
\begin{align}
\operatorname{prox}_{\delta}^\xset(\xref)
:=
\underset{(x,\delta)\in\widehat{\xset}}{\arg\max};
\left[
\tilde{w}\, \delta-\|x-\xref\|_2^2
\right].
\label{eq:convexrob}
\end{align}
The linear margin term rewards robustness, whereas the proximal term penalizes deviation from $\xref$, with $\tilde{w}>0$ balancing the two objectives.
Since the objective is concave and $\widehat{\xset}$ is convex, \eqref{eq:convexrob} is likewise a convex program.

\subsection{Feasibility Projection: A Shortest-Path Problem} \label{sec:nonconvexproj}
Given a trajectory $\xref$, we seek its projection onto the feasible set $\fset := \fset(\delta_\phi=0)$.
\begin{problem}[Euclidean Projection onto $\fset$] \label{prob:projection}
\begin{align}
    \Pi_\fset(\xref) = \underset{x \in \fset}{\arg\min} ||x - \xref||_2^2.
\end{align}
\end{problem}
Since $\fset$ is nonconvex, its minimizer may not be unique, however, it always exists within a finite union of convex sets. The minimization can therefore be decomposed over paths,
\begin{align}\label{eq:min-union}
    \min_{x \in \fset} \|x-\xref\|_2^2
    = \min_{\pi \in G_\fset} \left(\min_{x \in \fset_\pi \cap \tset_\pi} \|x-\xref\|_2^2\right), \nonumber
\end{align}
and if $\pi^*$ attains the outer and $x^*$ the corresponding inner minimum, then $x^*$ solves Problem~\ref{prob:projection}.
This confines the combinatorial complexity to the choice of path, while each corresponding inner subproblem is convex and admits a unique solution.
Moreover, the inner problem separates into spatial and temporal projection problems for $\fset_\pi$ and $\tset_\pi$, respectively.
Accordingly, we consider the following \emph{Shortest-Path Problem} (SPP) over $G_\fset$,
\begin{align}
    &\min_{\pi \in G_\fset}\sum_{(m,k) \in \pi} \Big[\underbrace{\min_{x \in \fset_{m,k}} ||x_m - \xref_m||_2^2}_{c^\fset_{m,k}} + \underbrace{\min_{x \in \tset_{m,k}} ||x - \xref||_2^2}_{c^\tset_{m,k}}\Big], \nonumber
\end{align}
with vertex costs $c_{m,k} = c^\fset_{m,k} + c^\tset_{m,k}$.
For brevity, we used $k=(o,\lambda,\gamma)$ and wrote $\fset_{m,k}:=\fset_{m,o,\lambda}$ and $\tset_{m,k}:=\tset_{m,\gamma}$.
The local structure of $\fset_\pi$ admits an exact decomposition into additive path cost. Each cost $c^\fset_{m,k}$ is a spatial projection of a single segment $m$ onto a single set $\fset_{m,k}$, computed once per vertex.
The temporal constraints $\tset_{m,k}$ couple preceding segment durations and projecting onto each set individually may under- or over-approximate the cost for projecting onto all constraints simultaneously. Hence, temporal costs $c_{m,k}^\tset$ are heuristic costs which become exact as $\|T - T'\|\rightarrow 0$.

We address Problem~\ref{prob:projection} by first solving the SPP over $G_\fset$ to determine $\pi^*$, after which projection onto the corresponding feasible sets yields $x^*=\Pi_{\fset_{\pi^*}\cap\tset_{\pi^*}}(\xref)\in\fset$.

\subsection{Proximal Robustness: A Widest-Path Problem} \label{sec:proxmax}
Given a trajectory $\xref$, the proximal robustness maximization problem is defined as follows.
\begin{problem}[Proximal Robustness]\label{prob:robustness}
\begin{align}
    \operatorname{prox}_{\delta_\phi}^\fset (\xref) = \underset{(x,\delta_\phi)\in \widehat{\fset}}{\arg\max} \left[\tilde{w}\,\delta_\phi - \|x - \xref\|_2^2  \right],
\end{align}
\end{problem}
\noindent where the proximal term anchors the solution to $\xref$ and the parameter $\tilde{w}$ trades margin against deviation.

Similar to \eqref{eq:min-union}, we can perform the maximization over the union $\fset$ for every path $\pi \in G_\fset$ separately,
\begin{align}\label{eq:max-union}
    \operatorname{prox}_{\delta_\phi}^\fset (\xref) = \max_{\pi \in G_\fset}\left(\max_{(x, \delta_\phi) \in \widehat{\fset}_\pi\cap\widehat{\tset}_\pi}\tilde{w}\,\delta_\phi - \|x-\xref\|_2^2\right). \nonumber
\end{align}
Generally, for a path $\pi$, the overall margin $\delta_\phi\in\{\rho_\phi,\theta_\phi\}$ is limited by the smallest margin among its components $(m,k)\in \pi$. This naturally motivates a $\min$-$\max$ objective,
\begin{align}
    \max_{\pi \in G_\fset} \min_{(m,k) \in \pi} \underbrace{\max_{(x,\delta_\phi) \in \widehat{\fset}_{m,k}\cap\widehat{\tset}_{m,k}}\Big[\tilde{w}\,\delta_\phi - \|x_m-\xref_m\|_2^2\Big]}_{r_{m,k}}, \nonumber
\end{align}
which constitutes a \emph{Widest-Path Problem (WPP)} over $G_\fset$ with vertex weight $r_{m,k}$, yielding the path with the largest (proximal) bottleneck.
Note that we have relaxed the proximal term $\|x_m-\xref_m\|_2^2 \leq \|x-\xref\|_2^2$ to allow for this reformulation and use heuristic lower-bound cost to decouple temporal constraints. The graph-search is therefore biased towards the path with the largest bottleneck margin but becomes \emph{exact} as $x^*$ approaches feasibility $\|x^* - \xref\|_2 \to 0$. Both properties are favorable in our nonconvex setting: early iterations are biased towards wide corridors, while exactness near convergence guarantees optimality.

Each vertex problem $r_{m,k}$ is a strictly concave, uniquely solvable problem as in \eqref{eq:convexrob} that is computed once per segment-set pair.
Let $m^*$ denote the bottleneck segment of $\pi^*$ and $\delta_\phi^*\in\{\rho_\phi^*,\theta_\phi^*\}$ its margin. Projecting each segment onto its superlevel set,
$x^* = \Pi_{\fset_{\pi^*}(\rho_\phi^*)\cap\tset_{\pi^*}(\theta_\phi^*)}(\xref)$, then yields the solution attaining $\delta_\phi^*$ on every segment.

\section{ADMM SOLVER}

\begin{figure*}[t]
  \centering
  \includegraphics{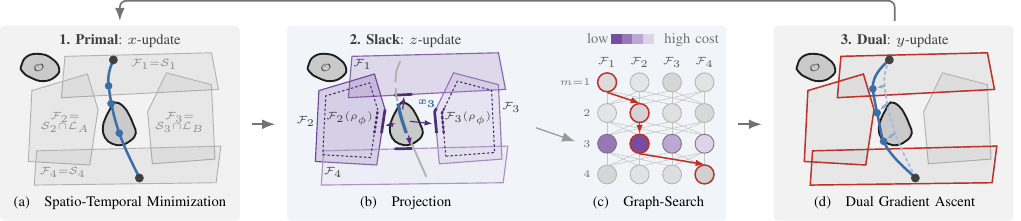}
  \caption{ADMM Iterations: (a) The primal update solves an unconstrained spatio-temporal minimization, yielding a continuous trajectory $x$ (blue). The slack update requires projecting onto the nonconvex feasible set, $\Pi_\fset(x)$, and robustness maximization, $\operatorname{prox}^\fset_{\delta_\phi}(x)$. Both are resolved by first projecting each segment $x_m$ onto each set $\fset_{m,k}$ (b) and solving a Pareto-optimal graph-search problem with associated costs (c). The dual gradient ascent step gradually enforces consensus between primal and slack variables. It acts as a cost update that shifts the primal solution towards feasibility as depicted in (d). }\label{fig:admm}
  \vspace{-5mm}
\end{figure*}

Our approach rests on one main observation: the difficulty of Problem~\ref{prob:general} lies in the coupling of continuous variables and discrete decisions, while each is tractable in isolation. We therefore propose a nonconvex ADMM-based solver that separates the smooth from the nonsmooth parts and decomposes the overall problem into subproblems that are well-understood and can be solved efficiently.

\subsection{Problem}
\newcommand{\bset}{\mathcal{B}}
We solve the following numerical problem with spatio-temporal parameters $x = (\mathbf{b}, \mathbf{T})$ and the graph-based feasibility set $\fset$ introduced in Sec.~\ref{sec:graphs}
\begin{subequations} \label{eq:numproblem}
\begin{align}
    \min_{x \in \mathcal{C}, \delta_\phi \geq0} & \quad J(x) - w \delta_\phi \\
    \text{s.t.} & \quad g(x) \in \bset, \\
            & \quad x \in \fset(\delta_\phi).
\end{align}
\end{subequations}
where $g(x)$ represents the smooth nonlinear mapping to the lifted parameters $\xl = (\mathbf{b}, \mathbf{\tilde{T}})$ as defined in Sec \ref{sec:spatiotempoptim}. The bounded set $\bset := \{\xl: b_{m,j}^{(r)} \subseteq \tilde{T}_{m,r}\mathcal{P}^r,\, \forall j, r, m\}$ limits derivatives through convex bounds $\pset^r$ and $\mathcal{C}$ enforces continuity which is handled implicitly by converting to the knot-state representation when optimizing over $x$ directly.

\subsection{Equivalent Consensus Problem}
Problem~\eqref{eq:numproblem} can be cast into a form amenable to ADMM~\cite{boyd2011distributed} by absorbing all set constraints into the objective via indicator functions, where $\mathbb{I}_\mathcal{Z}(z)$ equals zero if $z\in \mathcal{Z}$ and $+\infty$ otherwise. We introduce auxiliary slack variables $z = (z_R, z_\fset, z_\bset)^\top$ for robustness, feasibility ($\delta_\phi = 0$) and derivative bounds to separate smooth from nonsmooth terms:
\begin{align}
    \min_{x \in \mathcal{C},z, \delta_\phi}\, & \overbrace{J(x)}^\text{smooth} - \overbrace{w \delta_\phi+\mathbb{I}_{\fset(\delta_\phi)}(z_R)+\mathbb{I}_{\fset}(z_\fset) + \mathbb{I}_{B}(z_B)}^\text{nonsmooth} \nonumber\\
    \text{s.t.}\quad &\mathbf{g}(x) = z, \label{eq:consensusproblem}
\end{align}
 where $\mathbf{g}(x) = [x, x, g(x)]^\top$ enforces consistency between the primal and each of the slack variables.

\subsubsection{Augmented Lagrangian}
\newcommand{\La}{L_A}
Building on this consensus formulation, we present the standard Augmented Lagrangian (AL) of the equality-constrained problem in~\eqref{eq:consensusproblem}:
\begin{align}
    \La(x, z, y) =\: &J(x) - w \delta_\phi+\mathbb{I}_{\fset(\delta_\phi)}(z_R)
    +\mathbb{I}_{\fset}(z_\fset) + \mathbb{I}_{\bset}(z_\bset) \nonumber\\& + \frac{\beta}{2} \|\mathbf{g}(x) - z + y\|_2^2,
\end{align}
where $y$ are the scaled Lagrange multipliers and $\beta >0$ a scalar penalty parameter \cite{boyd2011distributed}.

\subsection{Alternating-Direction Method of Multipliers (ADMM)}
The \emph{Method of Multipliers} minimizes the augmented Lagrangian w.r.t. $(x,z)$ followed by a dual-ascent step on $y$.
Alternating minimization over $x$ and $z$ yields the three-step ADMM iteration:
\begin{subequations}
    \begin{align}
        \text{primal:}\quad x^{k+1} &= \arg\min_{x \in \mathcal{C}} \La(x, z^k, y^k), \\
        \text{slack:}\quad z^{k+1} &= \arg\min_{z} \La(x^{k+1}, z, y^k), \\
        \text{dual:}\quad y^{k+1} &= y^k + g(x^{k+1}) - z^{k+1},
    \end{align}
\end{subequations}
the last step of which is a simple gradient-ascent update on the Lagrange multiplier.
The primal and slack minimization steps are detailed in the following.

\begin{table*}[htb!]
    \caption{Comparison of our ADMM solver against GCS+STL~\cite{chen2026signal} and MIQP/MILP~\cite{verhagen2024temporally} in several scenarios.
    A `$-$' denotes a timeout after no solution is found at $10^{-3}$ tolerance within $1200$ seconds.
    Our ADMM method obtains maximal spatial robustness and close to optimal temporal robustness at a reduced computational time compared to the MIQP/MILP. We outperform the GCS+STL benchmark~\cite{chen2026signal} on all but one scenario while simultaneously being able to maximize the robustness metrics.}
    \label{tab:comparison}
    \centering
    \footnotesize %

    \begin{tabular*}{\textwidth}{@{\extracolsep{\fill}}l*{12}{c}@{}}
        \toprule
         & \multicolumn{3}{c}{ADMM 5$^{\text{th}}$ order}
         & \multicolumn{3}{c}{GCS+STL 5$^{\text{th}}$ order}
         & \multicolumn{3}{c}{STL MIQP 5$^{\text{th}}$ order}
         & \multicolumn{3}{c}{\gc{STL MILP 2$^{\text{nd}}$ order}} \\
        \cmidrule(lr){2-4}
        \cmidrule(lr){5-7}
        \cmidrule(lr){8-10}
        \cmidrule(lr){11-13}
        Scenario
         & $t$ [s] & $\theta$ & $\rho$
         & $t$ [s] & $\theta$ & $\rho$
         & $t$ [s] & $\theta$ & $\rho$
         & \gc{$t$ [s]} & \gc{$\theta$} & \gc{$\rho$} \\
        \midrule
        \texttt{stlcg}
         & \textbf{0.74} & 3.11 & \textbf{0.20}
         & 11.78 & 0.00 & 0.00
         & 22.30 & 4.10 & \textbf{0.20}
         & \gc{\textbf{0.34}} & \gc{\textbf{4.95}} & \gc{\textbf{0.20}} \\
        \texttt{puzzle-1}
         & 2.88 & \textbf{12.96} & \textbf{0.50}
         & \textbf{0.48} & 0.00 & 0.00
         & -- & -- & --
         & \gc{--} & \gc{--} & \gc{--} \\
        \texttt{rover}
         & \textbf{5.99} & 13.21 & \textbf{0.70}
         & 349.34 & 0.01 & 0.00
         & -- & -- & --
         & \gc{100.08} & \gc{\textbf{19.30}} & \gc{\textbf{0.70}} \\
        \texttt{either-or}
         & \textbf{1.42} & \textbf{7.78} & \textbf{0.55}
         & 65.53 & 0.29 & 0.13
         & -- & -- & --
         & \gc{--} & \gc{--} & \gc{--} \\
        \texttt{deliver}
         & \textbf{7.54} & 2.51 & \textbf{0.50}
         & 23.92 & 1.00 & 0.09
         & -- & -- & --
         & \gc{29.29} & \gc{\textbf{3.78}} & \gc{\textbf{0.50}} \\
        \bottomrule
    \end{tabular*}
    \vspace{-4mm}
\end{table*}

\begin{figure*}[t!]
    \centering
    \includegraphics[width=\textwidth]{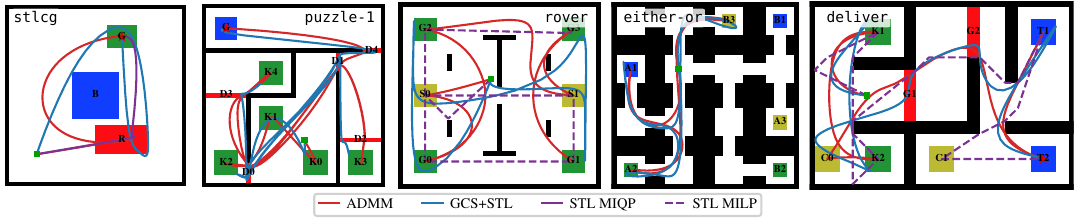}
    \vspace{-10mm}
    \label{fig:stl_scenarios}
\end{figure*}

\subsubsection{Primal Update}
The primal update takes the general form of a \emph{spatio-temporal minimization} as in Sec.~\ref{sec:spatiotempoptim},
\begin{align}
   x^{k+1} = \arg\min_{x \in \mathcal{C}} J(x) + \tfrac{\beta}{2}||\mathbf{g}(x) - c^k||_2^2,
\end{align}
with $c^k = z^k - y^k$. We minimize $x$ via the bijective mapping to $(\mathbf{D},\mathbf{T})$ and performing alternating minimization \cite{wang2020alternating}.

\subsubsection{Slack Update}
Minimizing $\La$ w.r.t. $z$ decomposes into individual subproblems for each slack variable. With $z^k = \mathbf{g}(x^{k+1}) + y^k$, we obtain a \emph{proximal robustness} problem for $z_R$ and \emph{Euclidean projection} problems for $z_\fset$ and $z_B$,
\begin{align}
z_R^{k+1}
&= \underset{(z_R, \delta_\phi) \in \widehat{\fset}}{\arg\max}\;
        \tilde{w}\,\delta_\phi
        -\lVert z_R-z_R^{k}\rVert_2^2
&&= \operatorname{prox}_{\delta_\phi}^\fset
    \!\left(z_R^{k}\right),
\nonumber
\\
z_\fset^{k+1}
&= \underset{z_\fset \in \fset}{\arg\min}\;
        \lVert z_\fset-z_\fset^{k}\rVert_2^2
&&= \Pi_\fset\!\left(z_\fset^{k}\right),
\nonumber
\\
z_\bset^{k+1}
&= \underset{z_\bset \in \bset}{\arg\min}\;
        \lVert z_\bset-z_\bset^{k}\rVert_2^2
&&= \Pi_\bset\!\left(z_\bset^{k}\right).
\nonumber
\end{align}
where the convex projection $\Pi_\bset\!\left(z_\bset^{k}\right)$ can be performed separately for each segment.
For $\operatorname{prox}_{\delta_\phi}^\fset(z_R^{k})$ and $\Pi_\fset(z_\fset^{k})$, we introduced efficient graphs-search surrogates in Sec.~\ref{sec:nonconvexproj} and~\ref{sec:proxmax}.
As both operators induce discrete choices on the \emph{same} feasibility graph, we search for a single path optimizing both objectives.
We first solve a set of small convex programs to obtain graph-weights $(c_{m,k}, r_{m,k})$ and pose the joint problem as a \emph{Pareto-optimal path search} problem,
\begin{align}
    \min_{\pi \in G_F}& \Biggl[\underbrace{\sum\nolimits_{(m,k)\in \pi} c_{m,k}}_{\textnormal{shortest-path:}\: C(\pi)} \underbrace{- \min_{(m,k)\in \pi} r_{m,k}}_{\textnormal{widest-path:} \: -R(\pi)}\Biggr].
\end{align}
The objective $f(\pi) = C(\pi) - R(\pi)$ is monotone decreasing in $R$ and monotone increasing in $C$ so any minimizer must be \textit{Pareto-optimal} with respect to $(C,R)$.
Computing the Pareto set on a directed acyclic graph can be performed efficiently, through $M$ dynamic programming iterations.

\subsection{Convergence}
Unlike the convex case, there exists no global convergence guarantee for ADMM in general nonconvex settings. Nevertheless, ADMM-based methods have been successfully applied to a broad class of nonconvex problems, including formulations involving unions of convex sets \cite{takapoui2020simple, pries2026admm}, nonlinear coupling constraints \cite{wang2021nonconvex}, and heuristics \cite{boyd2011distributed} as considered here.
In line with these and other methods like GCS \cite{marcucci2024shortest}, we rely on empirical validation and demonstrate fast and reliable convergence across challenging problems.

\section{RESULTS}
\label{sec:validation}

\begin{figure}[b!]
    \vspace{-5mm}
    \centering
    \includegraphics[width=0.48\textwidth]{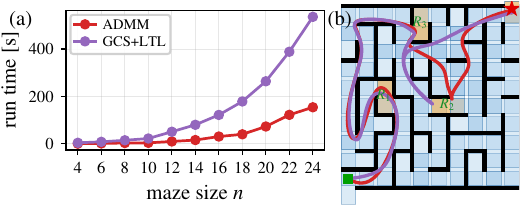}
    \caption{(a) Runtime for the GCS formulation and our ADMM approach as a function of the maze size. Results are averaged over five runs, each maze is $n \times n$. (b) A single $10\times10$ maze.}
    \label{fig:ltl_maze_timing}
    \vspace{-3mm}
\end{figure}

\subsection{Scaling}
We highlight the computational speed-up achieved by the explicit separation of the union of convex sets and the graph search.
We generate random mazes with the temporal logic specification $\phi_{\texttt{maze}} = \bigwedge_{i=1}^{3}F(x\in R_i) \land GF(x \in \text{Goal})$, requiring the visit of three random regions in the maze and ending up at the goal.
\cref{fig:ltl_maze_timing} compares our ADMM solver against GCS+LTL~\cite{kurtz2023temporal} which solves the product graph formulation as a single second-order cone program.
Our solver achieves a $4.7\times$ average speed-up.

\subsection{Signal Temporal Logic Benchmarks}
We compare STL motion planning on a set of (adapted) benchmark scenarios from~\cite{chen2026signal,sun2022multi} against two baselines: a GCS shortest-path product formulation of a timed automaton~\cite{chen2026signal} with clock constraints on the edges, and a globally optimal MIQP encoding~\cite{verhagen2024temporally}.
We additionally compare against a piecewise-linear version of the mixed-integer formulation.
\begingroup
\begin{equation*}
\resizebox{\linewidth}{!}{$\displaystyle
\begin{aligned}
    \phi_{\texttt{stlcg}} &= F_{[0,15]}G_{[0,5]}(x \in R) \land F_{[0,15]}G_{[0,5]}(x \in G) \land G_{[0,20]}(x \notin B) \\
    \phi_{\texttt{puzzle}} &= F_{[0,T]}(x \in G) \land \bigwedge_{i=0}^{4} (x \notin D_i) \mathcal{U}_{[0,T]}(x \in K_i) \\
    \phi_{\texttt{rover}} &= \bigwedge_{i=0}^{3} F_{[0,T]}(x\in G_i) \land \bigwedge_{i=0}^{1} (x \notin G_i \land x \notin G_{i+2})\mathcal{U}_{[0,T]} (x \in S_i) \\
    \phi_{\texttt{either}} &= \bigwedge_{i=1}^{3} F_{[0,T]}(x \in A_i) \land F_{[0,T]}(x \in B_i) \\
    \phi_{\texttt{deliver}} &= \bigvee_{i=0}^{1} F_{[0,20]}(x\in C_i) \land \bigwedge_{i=1}^{2} (x \notin G_i) \mathcal{U}_{[2,10]} (x \in K_i) \land \\ &F_{[10,26]}(x \in T_1) \land F_{[20,36]}(x\in T_2)
\end{aligned}$}
\end{equation*}
\endgroup

\cref{tab:comparison} reports trajectories, timing, and robustness metrics.
The MIQP (and even the simplified 2$^{\text{nd}}$-order MILP) finds the global optimum but times out as scenarios grow; GCS+STL scales better but, as a shortest-path formulation, cannot maximize robustness.
Our method outperforms GCS+STL w.r.t. speed and achieves optimal spatial- and near-optimal temporal robustness margins.
In \texttt{stlcg}, both our method and GCS+STL settle on a suboptimal looping solution, indicating that neither is guaranteed to recover the global optimum.

\subsection{Robust Satisfaction}
\begin{figure}[t!]
    \centering
    \includegraphics[width=0.48\textwidth]{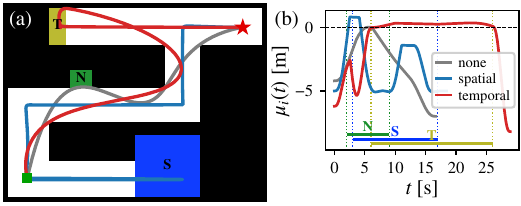}
    \caption{(a) Distinct trajectories for different robustness types; shortest path by visiting $N$, spatially robust by visiting $S$, and temporally robust by visiting $T$. (b) The degree of satisfaction of the chosen predicate $\mu_i(t)$ over time; spatially robust maximizing the value, temporally robust maximizing the interval.}
    \label{fig:stl_robustness}
    \vspace{-5mm}
\end{figure}

We show that our ADMM approach finds different traces satisfying a disjunctive STL specification depending on which robustness metric $\delta \in \{\rho,\theta\}$ is maximized. Consider
\begin{equation*}
    \phi_{\texttt{robust}} = F_{[2,9]}(x \in N) \lor F_{[3,17]}(x \in S) \lor F_{[6,26]}(x\in T)
\end{equation*}
requiring a visit to $N$, $S$, or $T$ within their respective intervals.
\cref{fig:stl_robustness} shows three runs, each maximal under its own robustness criterion: \emph{none} converges to the shortest path, \emph{spatial robustness} $\rho_\phi$ to visiting the largest region, and \emph{temporal robustness} $\theta_\phi$ to $T$, whose wide interval $[6,26]$ affords the greatest temporal margin despite being farthest from the start.
This shows our approach yields fundamentally different results than rounding a non-robust GCS-STL solution and using it to warm-start an NLP for the robust formulation.

\subsection{Large-Scale Rescue Scenario}
We lastly show how our ADMM method scales to large and complex scenarios, beyond the capabilities of current STL encodings~\cite{sun2022multi,verhagen2024temporally,chen2026signal}.
We consider a drone in an underground cave of the DARPA Subterranean Challenge~\cite{subt}.
Using the given cave geometry, we construct a convex set decomposition using \text{IRIS-ZO}~\cite{werner2024faster}.
The caves have locations of rescue Randys and accompanying phones. We specify that two low-critical patients need to be found after contact can be made with their phones and that a final high-critical patient should be visited directly. Eventually the robot should return to the base area to report
\begin{multline*}
    \phi_{\texttt{cave}} = \bigwedge_{i\in\{1,2\}} \bigl((x \notin \text{Randy}_i) \mathcal{U}_{[0,0.75 t_f]} (x \in \text{Phone}_i) \land \\ F_{[0,0.75t_f]}G_{[0,5]}(x \in \text{Randy}_i) \bigr) \land \\
    F_{[0,0.3t_f]}G_{[0,5]}(x \in \text{Randy}_3) \land F_{[0.75 t_f,t_f]}G_{[0,5]}(x \in \text{Base}).
\end{multline*}
The resulting trajectory is shown in \cref{fig:1}.
The convex decomposition of such a large and complex scene results in 324 boxes and $\phi_{\text{cave}}$ results in 409 TA states.
The rolled out product graph contains 28 million states when expanded over 212 segments.
The ADMM algorithm solves this large-scale problem in 257 seconds while the SOCP formulation of ~\cite{chen2026signal} exceeds the available memory.

\section{CONCLUSIONS}
\label{sec:conclusions}
This work presented a unified framework for continuous-time motion planning under collision-avoidance and temporal-logic constraints with explicit robustness maximization. Its core contribution is a joint feasibility graph that captures the discrete structure of safety and logic constraints and enables efficient graph-based subproblems within an ADMM splitting scheme. Across established benchmarks, the proposed method achieves state-of-the-art performance and converges reliably, while scaling to substantially larger motion-planning problems than those considered by existing approaches.

\bibliographystyle{IEEEtran}
\bibliography{references}

\end{document}